%% file: main_arxiv.tex
\PassOptionsToPackage{dvipsnames}{xcolor}
\documentclass[11pt,letterpaper]{style}

\usepackage[numbers]{natbib}
\usepackage{graphicx}
\usepackage{booktabs}
\usepackage{amsmath,amsfonts,amssymb}
\usepackage{cleveref}
\usepackage{subcaption}
\usepackage{wrapfig}
\usepackage{multirow}
\usepackage{colortbl}
\usepackage{listings}
\usepackage{xparse}
\usepackage{fontawesome5}
\usepackage{bxcoloremoji}
\usepackage{float}
\usepackage{placeins}
\usepackage{threeparttable}

\graphicspath{{./}{fig/}{figures/}{plot/}{pdf/}{table/}}
\usepackage{amsthm}
\usepackage{tcolorbox}
\usepackage{svg}
\tcbuselibrary{skins,breakable}
\tcbuselibrary{listingsutf8}
\usepackage{titletoc}
\microtypesetup{nopatch=footnote}
\usepackage{setspace}
\usepackage{pifont}
\usepackage{mathtools}
\usepackage{enumitem}
\usepackage{arydshln}
\usepackage{bbm}
\usepackage{lineno}
\usepackage{makecell}
\usepackage{adjustbox}
\usepackage{algorithm}
\usepackage{algorithmic}
\usepackage{caption}
\usepackage{hyperref}
\usepackage{bbding}
\usepackage{xspace}

\definecolor{mygray}{gray}{0.9}
\definecolor{syncol}{RGB}{243,246,249}
\definecolor{wildcol}{RGB}{215,240,235}
\definecolor{drop1}{RGB}{180,225,220}
\definecolor{drop2}{RGB}{150,210,200}
\definecolor{drop3}{RGB}{120,195,185}
\definecolor{drop4}{RGB}{95,180,170}
\definecolor{drop5}{RGB}{65,160,150}
\definecolor{lightblue}{RGB}{26,82,249}

\definecolor{myblue1}{HTML}{0171DC}
\definecolor{myblue2}{HTML}{013978}

\NewDocumentEnvironment{minted}{O{} m +b}{%
}{}

\newcommand{\equal}{\textsuperscript{*}}

\renewcommand\Authfont{\centering\normalfont\bfseries\fontsize{11}{15}\selectfont}
\renewcommand\Affilfont{\centering\normalfont\fontsize{10}{15}\selectfont}

\title{Steering the Teacher: Improving On-Policy Self-Distillation for Thinking Models}
\runningtitle{Steering the Teacher: Improving On-Policy Self-Distillation for Thinking Models}

\author{%
    {\Authfont
    \textbf{Tiejin Chen}\textsuperscript{*1} \quad
    \textbf{Xiaoou Liu}\textsuperscript{*1} \quad
    \textbf{Hanyang Chen}\textsuperscript{1} \quad
    \textbf{Hua Wei}\textsuperscript{1}
    }\\
    {\Affilfont
    \textsuperscript{1} Arizona State University \quad
    \texttt{\{tchen169, xiaoouli, hchen478, hua.wei\}@asu.edu} \\
    \textsuperscript{*} Equal Contribution
    }
}

\input{wordcard-style}

\begin{document}


\begin{abstract}
On-policy self-distillation (OPSD) trains a student on its own reasoning trajectories using token-level supervision from a reference-conditioned teacher initialized from the same model. However, access to the reference can induce teacher behaviors that undermine independent reasoning in thinking models, leading to reference hallucinations, where the student cites an unavailable reference to justify reasoning steps, and reconsideration suppression, where the teacher discourages verification and revision at uncertain steps. We introduce \textbf{Steered-Teacher OPSD (ST-OPSD)} to jointly address these failure modes through activation steering. ST-OPSD extracts two directions from model-generated reasoning trajectories and applies them to the teacher's hidden states during distillation target computation, suppressing reference-invoking behavior and encouraging reconsideration with marker-family calibration. The student learns from the resulting teacher distributions without being steered. Experiments on AIME and HMMT show that ST-OPSD improves accuracy over vanilla OPSD and other interventions while restoring reconsideration and reference-hallucination rates to base-model levels. 
\end{abstract}

\newcommand{\TitleLinks}{%
\centering
    \vspace{8pt}
}

\maketitle


\input{1intro}

\input{2related}
\input{3method}
\input{4experiment}

\input{5conclusion}


\bibliographystyle{unsrtnat}
\bibliography{ref}


\appendix

\input{6appendix}

\end{document}

%% file: wordcard-style.tex
\usepackage{tcolorbox}
\definecolor{carddark}{RGB}{63,63,63}
\definecolor{cardlight}{RGB}{245,245,245}

\newtcolorbox{wordcard}[1]{
    title={#1},
    colback=cardlight,
    colframe=carddark,
    colbacktitle=carddark,
    coltitle=white,
    fonttitle=\bfseries,
    fontupper=\small,
    boxrule=0.7pt,
    arc=2mm,
    left=8pt,
    right=8pt,
    top=8pt,
    bottom=8pt,
    toptitle=4pt,
    bottomtitle=4pt,
    before upper={\setlength{\parindent}{0pt}\raggedright}
}

%% file: 1intro.tex
\section{Introduction}
\label{sec:introduction}

\begin{figure}[ht]
    \centering
    \includegraphics[width=0.8\linewidth]{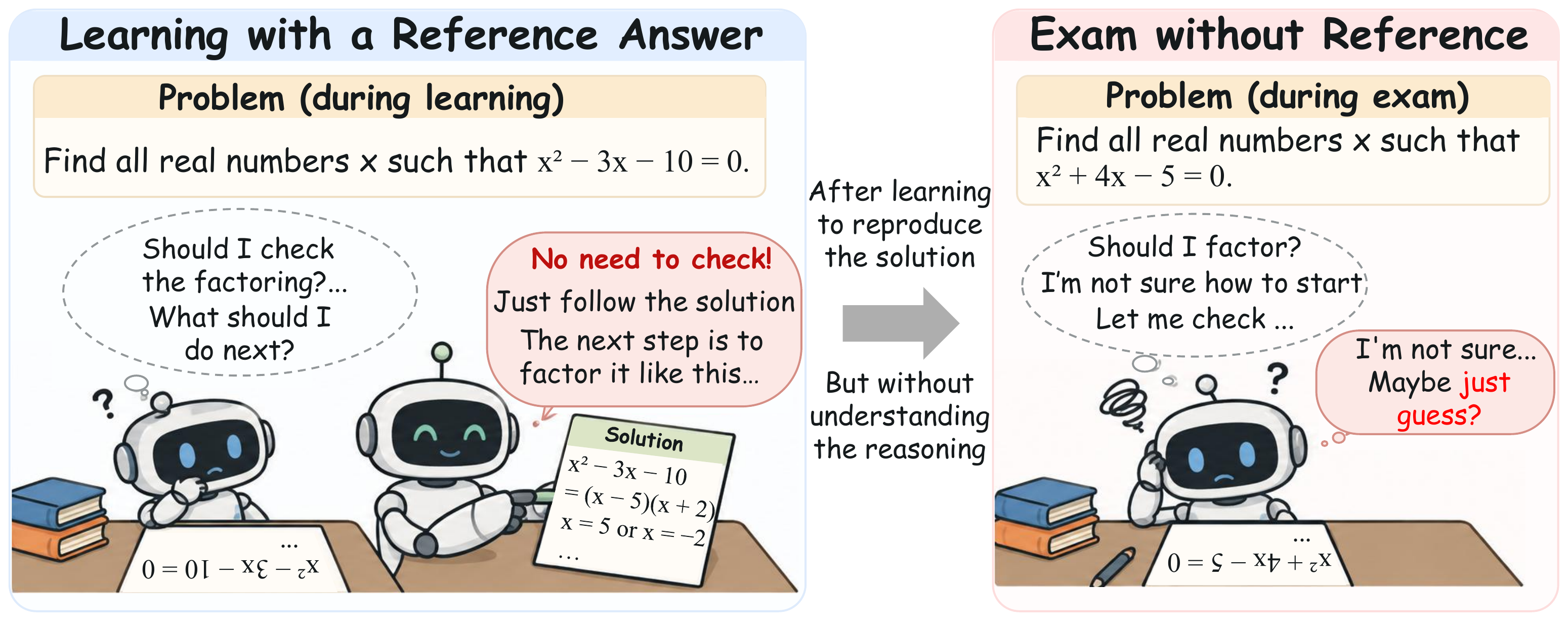}

    \caption{An analogy of reference-conditioned supervision in OPSD: a tutor prescribes solution steps and discourages verification, potentially leaving the student unable to reason independently on new problems without a reference answer.}
    \label{fig:intro}
\end{figure}

As illustrated in Figure~\ref{fig:intro}, a student tutors a classmate through a difficult math problem with access to the reference answer. When the classmate pauses to check a step or hesitates about what to do next, the tutor dismisses these efforts as unnecessary and tells them to follow the provided solution, pointing to the reference answer to prescribe the next step without explaining why it works. The classmate may learn to reproduce the solution without understanding the reasoning behind it or how to apply it to new problems. During an exam without the reference answer, they may lack an independent reasoning strategy, skip necessary checks, and ultimately resort to guessing.

Language models face a similar challenge in on-policy self-distillation (OPSD). A student generates reasoning from the question alone, while a teacher initialized from the same model additionally receives a reference solution and provides token-level supervision along the student's trajectory~\citep{zhao2026opsd}. This trains the student on contexts encountered during its own generation~\citep{agarwal2024gkd, gu2024minillm}, but reference-conditioned supervision may also transfer behaviors that undermine independent reasoning.

Specifically, there are two teacher behaviors that contribute to performance degradation in thinking models under OPSD. First, the teacher invokes privileged information through phrases such as \textit{``The example solution uses...''}. The student learns to imitate these citations despite lacking access to the reference, producing \textbf{reference hallucinations}: reasoning steps justified by an unseen solution rather than independent derivation. These hallucinations manifest reference-induced shortcuts~\citep{shen2026purified, harne2026privilegedbiased}. Second, knowing the answer leads the teacher to assign low probabilities to reconsideration markers at uncertain branching points~\citep{kaur2026rethinking,peng2026diagnosing}. The resulting \textbf{reconsideration suppression} encourages the student to continue confidently~\citep{kaur2026rethinking, kim2026degrade}, shortening reasoning without corresponding accuracy gains and degrading performance under longer reasoning budgets~\citep{kaur2026rethinking}.

We introduce \textbf{Steered-Teacher OPSD (ST-OPSD)}, which addresses reference hallucinations and reconsideration suppression through activation steering, a technique that changes a model's behavior by adding directions to its hidden states~\citep{turner2023activation, panickssery2024caa}. We extract two directions from the model's own reasoning trajectories: a \emph{reference direction} that distinguishes reference-invoking steps from independent reasoning, and a \emph{reconsideration direction} that distinguishes sentence boundaries followed by reconsideration markers from other boundaries. 
While the teacher scores the students' on-policy trajectories, ST-OPSD subtracts the reference direction and adds the reconsideration direction to the teacher's hidden states. The resulting next-token distributions serve as distillation targets. The student itself is never steered and learns through parameter updates. Experiments on AIME and HMMT show that ST-OPSD improves accuracy over both vanilla OPSD and other baselines, while restoring the student model’s reconsideration frequency and reference-hallucination rate to base-model levels.

Overall, our contributions are:


\begin{itemize}
    \item We introduce ST-OPSD, a teacher-steering framework that
    uses two trajectory-derived directions to mitigate the reference
    hallucinations and reconsideration suppression in OPSD.

    \item We demonstrate that ST-OPSD improves accuracy on AIME
    and HMMT while restoring the student's reconsideration and
    reference-hallucination rates to base-model levels.

    \item We show that combining both steering directions achieves the strongest results. We further show that ST-OPSD achieves higher accuracy than random steering, teacher prompting, and direct student steering at inference time.
\end{itemize}

%% file: 2related.tex
\section{Related Work}

\paragraph{On-policy Self-Distillation.}
On-policy distillation provides token-level teacher supervision on student-generated prefixes, aligning the training states with the student's evolving policy~\citep{agarwal2024gkd, gu2024minillm}. On-policy self-distillation (OPSD) retains this supervision mechanism while using a privileged version of the model itself as the teacher. OPSD uses a privileged version of the model as the teacher, with prior work exploring different sources of privileged guidance, such as guidance or hints and teacher-student training designs~\citep{zhao2026opsd,shenfeld2026sdft,hubotter2026sdpo,penaloza2026privileged,chen2026dualopsd,zhang2026lopd}. The different designs expand how privileged guidance is obtained and maintained, while leaving its suitability as supervision for an unprivileged student an important concern. Though OPSD achieves better performance for instruction models, for thinking models, a reference-conditioned teacher can introduce two related risks. First, it can suppress hesitation, self-correction, and exploration, altering the student's deliberative behavior~\citep{kim2026degrade, kaur2026rethinking}. Second, it can favor answer-dependent shortcuts that are unavailable to the student at inference~\citep{harne2026privilegedbiased, shen2026purified}. Students trained under such supervision can inherit shorter reasoning traces and perform worse~\citep{kaur2026rethinking}. Meanwhile, reduced hesitation and reconsideration could also be associated with poorer out-of-distribution generalization~\citep{kim2026degrade}. Recent work has explored alternative ways of constructing supervision signals from privileged teachers. OGLS-SD~\citep{yang2026oglssd} contrasts teacher logits conditioned on successful and failed rollouts to reduce shared privileged-conditioning biases and improve outcome-discriminative guidance. Purified OPSD~\citep{shen2026purified} targets answer-dependent shortcuts by subtracting reference-only log probabilities from those of the question-and-reference-conditioned teacher to form a new training target. RLSD~\citep{yang2026rlsd} tries to address reference leakage at the objective level by using log-probability differences to reweight verifier-derived advantages. In contrast to these approaches, which modify supervision at the objective level, we intervene directly in the privileged teacher's hidden states to discourage reference-invoking behavior and preserve reconsideration during on-policy distillation. Our steering-vector approach also provides more direct and fine-grained control over these behaviors through separately adjustable steering strengths.

\paragraph{Steering Vectors.}
Steering vectors control language-model behavior by intervening in hidden representations~\citep{turner2023activation, li2023inference, panickssery2024caa, zou2023representation}. Empirical studies identify approximately linear directions associated with certain semantic and behavioral attributes~\citep{turner2023activation, zou2023representation}. A common construction takes the difference between mean activations from examples that differ in the target behavior~\citep{panickssery2024caa, zou2023representation}. Beyond behavioral directions, function vectors capture task-related representations~\citep{todd2024fv}, while task arithmetic explores the composition of task-specific updates in parameter space~\citep{ilharco2023editing}. Subsequent studies examine steering reliability~\citep{imbrasaite2025reliability}, refine vectors with sparse autoencoders~\citep{zhao2025saersv}, and develop conceptor-based activation transformations~\citep{valois2025conceptors}. Feedback controllers further adapt interventions during generation~\citep{subramani2025feedback}. Besides, applications extend to test-time optimization, visual understanding, and reasoning~\citep{tang2025ttsv, li2025textual, vyas2025thinking}. In this paper, instead of using the steering vector in inference, we jointly apply steering vectors to a privileged teacher during target computation to investigate how to reduce reference dependence and preserve hesitation during training state, shaping on-policy supervision for thinking models.

%% file: 3method.tex
\section{Method}
\label{sec:method}

\begin{figure*}
    \centering
    \includegraphics[width=\linewidth]{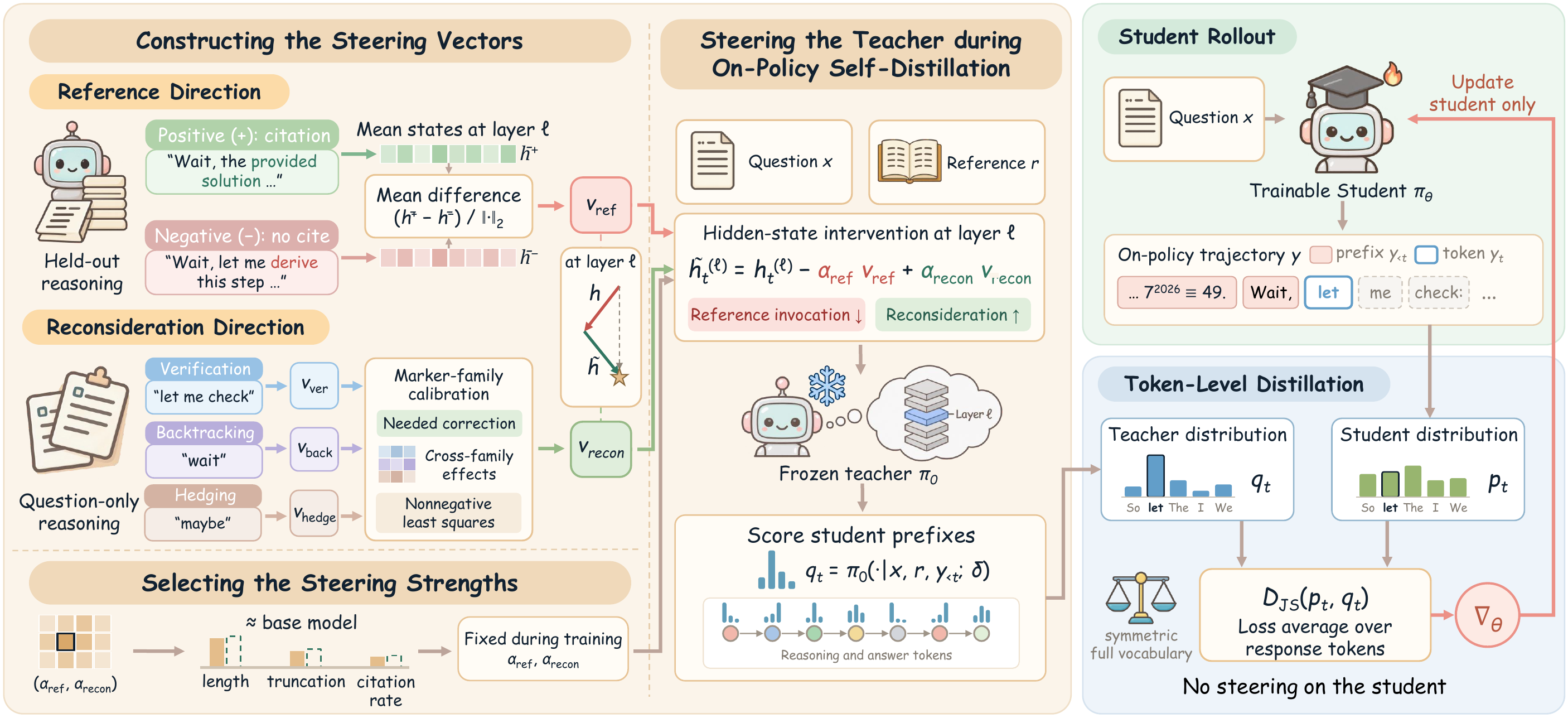}
    \caption{Overview of the ST-OPSD framework. We construct steering vectors and apply them to the teacher to reduce reference invocation and encourage reconsideration during on-policy self-distillation. The student learns from the steered teacher while generating reasoning from the question alone.}
    \label{fig:framework}
\end{figure*}

With the answer in the prompt, the teacher might choose to cite the reference directly and discourage reconsideration. Following the teacher, the student will also learn to cite a reference and skip reconsideration, which makes it harder to solve new problems independently when no reference answer is available. The key idea is to restore the teacher's support for independent reasoning and reconsideration. However, prompt instructions alone provide limited control over these behaviors. ST-OPSD therefore uses steering vectors to control these behaviors by directly modifying the teacher's hidden states~\cite{panickssery2024caa}. We will introduce how to construct the steering vector and how the steering vector will be used in OPSD next.

\subsection{Constructing the Steering Vectors}
\label{sec:method-vectors}

To promote these two behaviors in the teacher, we construct two steering directions. One supports independent reasoning by discouraging reference citation, while the other encourages reconsideration. Steering vector is motivated by the observation that behavioral tendencies can be encoded as approximately linear directions in a model's hidden-state space~\cite{panickssery2024caa}. Shifting hidden states along these directions can therefore promote or suppress the corresponding behaviors. A common way to estimate such a direction is to take the difference between the mean hidden states of examples that exhibit the target behavior and those that do not.

To construct the steering vector, we use the base model $\pi_0$ to generate reasoning trajectories on problems held out from OPSD training. These trajectories provide positive and negative examples of the behaviors we aim to control. Positive examples contain the target behavior and negative examples does not contain. How to obtain and select these examples will be described later.

For each positive or negative example $i$, we extract the hidden state $h_{i,t_i}^{(\ell)}$ at the output of layer $\ell$, where $t_i$ is the token position selected for that example. Let $\mathcal{I}^{+}$ and $\mathcal{I}^{-}$ denote the index sets of positive and negative examples, respectively. We then average the selected token states within each group to obtain a representative activation for each group, reducing the influence of variation across individual examples:
\begin{equation}
\label{eq:mean-states}
\begin{aligned}
    \bar{h}^{+}
    &= \frac{1}{|\mathcal{I}^{+}|}
       \sum_{i \in \mathcal{I}^{+}} h_{i,t_i}^{(\ell)} , \quad
    \bar{h}^{-}
    &= \frac{1}{|\mathcal{I}^{-}|}
       \sum_{i \in \mathcal{I}^{-}} h_{i,t_i}^{(\ell)}.
\end{aligned}
\end{equation}
Since the two groups differ in whether the target behavior is present~\cite{turner2023activation, li2023inference}, we use their mean difference to estimate the corresponding behavioral direction and normalize it to obtain a unit steering vector:
\begin{equation}
\label{eq:steering-direction}
    d = \bar{h}^{+} - \bar{h}^{-},
    \quad
    u = \frac{d}{\lVert d \rVert_2}.
\end{equation}
Normalization allows us to specify the steering direction and intervention strength separately.

\paragraph{Reference direction.}

To suppress the teacher's reference-citing behavior, we construct this direction from its own reasoning trajectories. Broadly suppressing reference-related activations could also weaken the teacher's use of privileged information. We therefore focus on reference citations after reconsideration.  Within these segments, we identify explicit reference citations through pattern matching against expressions such as ``example solution,'' ``provided solution,'' and ``in the example.'' Segments containing a matched expression serve as positive examples, while segments without a match serve as negative examples. After getting the positive and negative examples, we can obtain $v_{ref}$ using \cref{eq:steering-direction}.

\paragraph{Reconsideration direction.}
To restore the teacher's support for reconsideration, we use reasoning trajectories generated by the base model $\pi_0$ without the reference, which still contains the ability to reconsider. Following~\citet{kaur2026rethinking}, we identify reconsideration through three marker families, including verification (e.g., ``let me check''), backtracking (e.g., ``wait''), and hedging (e.g., ``maybe''). For each reconsideration family $f$, sentences beginning with one of its markers serve as positive examples. All families share a negative pool of sentences beginning with none of the markers in our inventory. Generic planning expressions such as ``let me think'' also signal deliberation, so we exclude sentences beginning with these expressions from the negative pool. Using \cref{eq:steering-direction}, we construct a unit direction $v_f$ for each family separately, so that their relative contributions are not determined by their sample counts. We then combine these directions with nonnegative weights calibrated to bring the teacher's marker probabilities closer to those of the base model without the reference. To determine the combination weights, we first measure the probability correction needed for the teacher to match the question-only base model on each marker family, using the same held-out reasoning prefixes. We then perturb the teacher along each family direction separately and measure the resulting probability changes across all three families. We fit nonnegative weights by least squares so that the weighted sum of these measured effects approximates the required correction. Finally, we normalize the weighted sum of the family directions to obtain $v_{\mathrm{recon}}$.

\subsection{Steering the Teacher during On-Policy Self-Distillation}
\label{sec:method-opsd}

We next use the two constructed steering directions to modify the teacher's supervision during OPSD. At each iteration, the student generates reasoning trajectories without access to the reference. The teacher then scores these trajectories with access to the reference solution, while its hidden states are steered to discourage reference citation and encourage reconsideration. The resulting token distributions serve as distillation targets for updating the student.

When the teacher scores a student-generated trajectory, we modify its hidden state at the output of layer $\ell$ as:

\begin{equation}
\label{eq:teacher-steering}
\widetilde{h}_t^{(\ell)}
=
h_t^{(\ell)}
-
\alpha_{\mathrm{ref}} v_{\mathrm{ref}}
+
\alpha_{\mathrm{recon}} v_{\mathrm{recon}},
\end{equation}
where $\alpha_{\mathrm{ref}}, \alpha_{\mathrm{recon}} \geq 0$ control the intervention strengths. Both directions point toward the behaviors represented by their positive examples. We therefore subtract $v_{\mathrm{ref}}$ to suppress reference citation and add $v_{\mathrm{recon}}$ to promote reconsideration. The intervention applies throughout the response, including positions that predict reasoning and final-answer tokens.

\paragraph{Selecting the steering strengths.}
To set the intervention strengths in \cref{eq:teacher-steering}, we evaluate candidate pairs on a small held-out set, separate from the data used for steering-vector construction and OPSD training. For each pair, we generate teacher responses with both directions active and compare them with responses from the base model on the same problems under the same token budget. Excessive reconsideration steering can produce overly long responses. We therefore select strengths that keep the teacher's average response length and truncation rate close to those of the base model. The selected strengths should also reduce reference citation after reconsideration to approximately the base model's rate as well. We measure this citation rate as the fraction of marker occurrences followed by a matched reference expression. The selected strengths remain fixed throughout OPSD training.

\paragraph{Learning from the steered teacher.}
With the steering strengths fixed, we train the student using token-level supervision from the steered teacher. At training step $k$, we sample a batch of problems $\{(x_i,r_i)\}_{i=1}^{B}$, where $x_i$ is the question and $r_i$ is its reference solution. The current student generates a response $y_i \sim \pi_{\theta_k}(\cdot \mid x_i)$ for each question. Let $\delta = -\alpha_{\mathrm{ref}}v_{\mathrm{ref}} + \alpha_{\mathrm{recon}}v_{\mathrm{recon}}$ denote the combined hidden-state perturbation. Along each sampled trajectory, the student and the steered teacher compute next-token distributions conditioned on the same response prefix,
\begin{equation}
\label{eq:distillation-distributions}
\begin{aligned}
p_{i,t}
&= \pi_{\theta}(\cdot \mid x_i, y_{i,<t}), \\
q_{i,t}
&= \pi_0(\cdot \mid x_i, r_i, y_{i,<t}; \delta),
\end{aligned}
\end{equation}
where $\delta$ specifies the intervention applied to the frozen teacher's hidden states according to \cref{eq:teacher-steering}. The resulting distribution $q_{i,t}$ serves as the distillation target for the student at each step.

We then train the student using the original OPSD, which matches these teacher distributions by minimizing the symmetric Jensen--Shannon divergence over the full vocabulary. With $\mu_{i,t} = \tfrac{1}{2}(p_{i,t}+q_{i,t})$, the divergence will be:
\begin{equation}
\label{eq:js-divergence}
D_{\mathrm{JS}}(p_{i,t},q_{i,t})
=
\frac{1}{2}\mathrm{KL}(p_{i,t}\,\|\,\mu_{i,t})
+
\frac{1}{2}\mathrm{KL}(q_{i,t}\,\|\,\mu_{i,t}).
\end{equation}
For responses of lengths $T_i = |y_i|$, the training objective averages this divergence over all generated reasoning and final-answer tokens,
\begin{equation}
\label{eq:st-opsd-objective}
\mathcal{L}_k(\theta)
=
\frac{
\displaystyle \sum_{i=1}^{B}\sum_{t=1}^{T_i}
D_{\mathrm{JS}}(p_{i,t},q_{i,t})
}{
\displaystyle \sum_{i=1}^{B} T_i
}.
\end{equation}
During each update, we hold the sampled trajectories and teacher distributions fixed and optimize only the student's parameters. The updated student then generates the next batch of on-policy trajectories. The student learns through these parameter updates and uses no steering during training or inference.

%% file: 4experiment.tex
\section{Experiments}

We conduct comprehensive experiments to evaluate the effectiveness of ST-OPSD, examine how teacher steering shapes student reasoning, and compare it with alternative interventions. Specifically, our experiments are organized around the following research questions:

RQ1: Does ST-OPSD improve student reasoning performance and behavior across models and training settings?


RQ2: How do individual steering components and alternative interventions affect the performance of ST-OPSD?


RQ3: How does ST-OPSD compare with other methods in computational efficiency?


\begin{table*}[t]
    \centering
    \small
    \setlength{\tabcolsep}{3pt}
    \resizebox{\textwidth}{!}{
    \begin{tabular}{ll*{9}{c}!{\vrule width 1pt}*{3}{c}}
        \toprule
        \multirow{2}{*}{Model}
        & \multirow{2}{*}{Method}
        & \multicolumn{3}{c}{AIME24}
        & \multicolumn{3}{c}{AIME25}
        & \multicolumn{3}{c!{\vrule width 1pt}}{HMMT25}
        & \multicolumn{3}{c}{Avg.} \\
        \cmidrule(lr){3-5}
        \cmidrule(lr){6-8}
        \cmidrule(lr){9-11}
        \cmidrule(lr){12-14}
        & & Acc $\uparrow$ & Hes./1k $\uparrow$ & Ref $\downarrow$
        & Acc $\uparrow$ & Hes./1k $\uparrow$ & Ref $\downarrow$
        & Acc $\uparrow$ & Hes./1k $\uparrow$ & Ref $\downarrow$
        & Acc $\uparrow$ & Hes./1k $\uparrow$ & Ref $\downarrow$ \\
        \midrule
        \multirow{5}{*}{Qwen3-4B}
        & Base
        & 70.8 & 12.9 & 10.8 & 65.8 & 13.1 & 15.8 & 41.7 & 15.1 & 14.2
        & \underline{59.4} & 13.7 & 13.6 \\
        & Vanilla OPSD
        & 69.2 & 13.4 & 59.2 & 56.7 & 12.9 & 68.3 & 31.7 & 15.4 & 72.5
        & 52.5 & 13.9 & 66.7 \\
        & Purified OPSD
        & 70.8 & 14.1 & 10.0 & 64.2 & 13.1 & 17.5 & 43.3 & 18.7 & 8.3
        & \underline{59.4} & \textbf{15.3} & \textbf{11.9} \\
        & Random SV
        & 69.2 & 13.2 & 56.7 & 49.2 & 13.3 & 78.3 & 31.7 & 16.6 & 79.2
        & 50.0 & 14.4 & 71.4 \\
        & ST-OPSD
        & 74.2 & 13.8 & 12.5 & 64.2 & 14.4 & 14.2 & 45.0 & 16.8 & 12.5
        & \textbf{61.1} & \underline{15.0} & \underline{13.1} \\

        \midrule
        \multirow{5}{*}{Qwen3-8B}
        & Base
        & 79.2 & 13.8 & 15.0 & 70.0 & 14.1 & 20.0 & 46.7 & 14.7 & 25.0 & \underline{65.3} & 14.2 & \underline{20.0} \\
        & Vanilla OPSD
        & 75.0 & 13.5 & 55.0 & 58.3 & 13.2 & 62.5 & 43.3 & 14.2 & 74.2 & 58.9 & 13.6 & 63.9 \\
        & Purified OPSD
        & 75.8 & 14.1 & 13.3 & 67.5 & 14.6 & 22.5 & 49.2 & 14.7 & 25.0 & 64.2 & \underline{14.5} & 20.3 \\
        & Random SV
        & 76.7 & 13.7 & 50.0 & 58.3 & 14.4 & 76.7 & 45.0 & 15.4 & 70.8 & 60.0 & 14.5 & 65.8 \\
        & ST-OPSD
        & 83.3 & 15.6 & 15.0 & 65.8 & 15.9 & 23.3 & 49.2 & 18.9 & 19.2 & \textbf{66.1} & \textbf{16.8} & \textbf{19.2} \\

        \midrule
        \multirow{5}{*}{OLMo-7B}
        & Base
        & 77.5 & 12.6 & 19.2 & 67.5 & 12.2 & 20.0 & 46.7 & 12.9 & 22.5
        & \textbf{63.9} & 12.6 & \underline{20.6} \\
        & Vanilla OPSD
        & 73.3 & 12.1 & 30.8 & 64.2 & 12.1 & 33.3 & 44.2 & 12.9 & 32.5
        & 60.6 & 12.4 & 32.2 \\
        & Purified OPSD
        & 75.0 & 12.3 & 21.7 & 68.3 & 12.4 & 26.7 & 44.2 & 12.7 & 27.5
        & 62.5 &  12.5 & 25.3 \\
        & Random SV
        & 70.8 & 13.6 & 23.3 & 65.0 & 13.1 & 31.7 & 36.7 & 12.2 & 31.7
        & 57.5 & \underline{13.0} & 28.9 \\
        & ST-OPSD
        & 75.8 & 13.8 & 16.7 & 68.3 & 13.6 & 8.3 & 45.8 & 14.2 & 15.0
        & \underline{63.3} & \textbf{13.9} & \textbf{13.3} \\
        \bottomrule
    \end{tabular}
    }
    \caption{Results on three mathematical reasoning benchmarks.
Acc. denotes avg@4 accuracy (\%), Hes./1k denotes the number of
hesitation markers per 1,000 generated tokens, and Ref. (\%) denotes
the percentage of responses containing reference citations.
Across all evaluated models, ST-OPSD improves average accuracy
over Vanilla OPSD while bringing hesitation frequency and
reference-citation rates closer to base-model levels.}
    \label{tab:main_results}
\end{table*}

\subsection{Experimental Setting}

\paragraph{Models}
To test our method comprehensively, we evaluate ST-OPSD on four different thinking models, including Qwen3-4B~\cite{yang2025qwen3}, Qwen3-8B~\cite{yang2025qwen3}, and OLMo-7B-Thinking~\cite{groeneveld2024olmo}. We evaluate models using the original OPSD method~\cite{zhao2026opsd}, where the student is updated through distillation, while the teacher remains frozen.

\begingroup
\setlength{\emergencystretch}{3em}

\paragraph{Datasets}
Following previous work~\cite{kaur2026rethinking}, we train all models on the filtered mathematical subset of OpenThoughts~\cite{guha2026openthoughts}. We evaluate on AIME 24, AIME 25~\cite{maa_aime}, and HMMT 25~\cite{balunovic2026matharena}. We reserve 64 OpenThoughts questions for constructing all steering vectors and exclude them from training. To assess the effect of the training dataset, we also train on Omni-Math~\cite{gao2025omni} and evaluate on the same benchmarks.

\par
\endgroup

\paragraph{Configuration}
We use the same LoRA~\cite{hu2022lora} settings with a rank of 64, learning rate $2\times10^{-5}$ as the previous paper~\cite{shen2026purified}. We also use an update step of 200 and a batch size of 32. The training rollout uses vLLM~\cite{kwon2023efficient} with a temperature of $1.0$, top-$p=1.0$, and a maximum generation length of 8,192 tokens. For evaluation, we sample four responses per problem at temperature $0.6$ and top-$p=0.95$, allowing up to 81,920 tokens per response, and we grade final answers using math\_verify~\cite{kydlicek2025mathverify} and report \texttt{avg@4}, the mean correctness over the four responses per problem and across problems, for accuracy. We will also report the frequency of hesitation words and mentions of answers in the prompt to demonstrate the change of our methods for the trained model. We compare ST-OPSD with Vanilla OPSD and Purified OPSD~\cite{shen2026purified} and use one Nvidia H100 for all experiments.

\paragraph{Steering details.}

\begin{table}[t]
    \centering
    \small
    \begin{tabular}{lccc}
        \toprule
        Model & $\alpha_{\mathrm{ref}}$
              & $\alpha_{\mathrm{recon}}$
              &  $\lVert v \rVert_2 /\lVert out \rVert_2$  \\
        \midrule
        Qwen3-4B & 2.29 & 9.50 & 12.7 \\
        Qwen3-8B & 4.25 & 12.8 & 8.9 \\
        OLMo-7B  & 0.80 & 3.50 & 18.7 \\
        \bottomrule
    \end{tabular}
    \caption{Steering coefficients and vector norms for each model.}
    \label{tab:steering_parameters}
    \vspace{-3mm}
\end{table}

For the steering vector, we apply all directions at layer 22 for each model we used. For each direction, we use a hold-out set of 64 questions and sample 2 generations for one question and obtain positive and negative samples. For the chosen $\alpha_{\mathrm{ref}}$ and $\alpha_{\mathrm{recon}}$, we provide the detailed parameters in \cref{tab:steering_parameters}. The 
ratio $\lVert v \rVert_2 /\lVert out \rVert_2$ indicates the magnitude of the steering vector relative to the output activation. For our selected steering vector, this ratio typically remains around 10\%.



\subsection{Effectiveness of ST-OPSD (RQ1)}

We evaluate whether correcting reference citation and reconsideration suppression in the teacher improves student learning. We compare ST-OPSD with the base model, Vanilla OPSD, Purified OPSD, and Random SV using Qwen3-4B, Qwen3-8B, and OLMo-7B on AIME24, AIME25, and HMMT25 using OpenThoughts as the training set. We report avg@4 accuracy, hesitation-marker frequency (Hes./1k), and reference-citation rate (Ref.) to examine both task performance and the behaviors learned during distillation. We present our results in \cref{tab:main_results} and we have the following observation:

\noindent \textbullet \quad ST-OPSD achieves higher average accuracy than Purified OPSD on all three models, alongside clear improvements over Vanilla OPSD. It also exceeds the base model on both Qwen models.  On OLMo-7B, ST-OPSD nearly matches the base model, recovering most of the accuracy lost under Vanilla OPSD. We hypothesize that the reference provides limited additional reasoning benefits to the OLMo teacher, restricting further gains beyond the base model. These results demonstrate that ST-OPSD improves the effectiveness of OPSD.

\noindent \textbullet \quad Vanilla OPSD makes the student cite the reference much more often. On HMMT25, over 70\% of responses from both Qwen models cite the reference, far above the base-model rates. ST-OPSD brings the average citation rate back to levels close to or below the base model on all three models. These results demonstrate that ST-OPSD effectively suppresses reference-citing behavior in the student.

\noindent \textbullet \quad ST-OPSD produces higher hesitation behavior than the base model on all three models, consistent with encouraging reconsideration during distillation. However, marker frequency alone does not determine accuracy. On Qwen3-4B, Purified OPSD has a higher hesitation frequency and a lower reference-citation rate than ST-OPSD, yet achieves lower accuracy. These results indicate that the behavioral metrics are useful diagnostics but do not fully capture the quality of the reasoning learned by the student.

\noindent \textbullet \quad Random SV that has the same norm as our final SV fails to reproduce the accuracy gains of ST-OPSD and can perform even worse than Vanilla OPSD. Thus, modifying the teacher's hidden states alone is insufficient to improve distillation. The advantage of ST-OPSD supports constructing steering directions that specifically target reference citation and reconsideration.

\subsection{Ablation Studies (RQ2)}

\label{sec:ablation}

In this section, we examine how the supervision design and training data affect ST-OPSD. All experiments use Qwen3-4B and are evaluated on AIME24, AIME25, and HMMT25.

\begin{figure*}[t]
    \centering
    \captionsetup[subfigure]{
        labelformat=parens,
        labelsep=space
    }
    \renewcommand{\thesubfigure}{\alph{subfigure}}

    \begin{subfigure}[t]{0.32\textwidth}
        \centering
        \includegraphics[width=\linewidth]
            {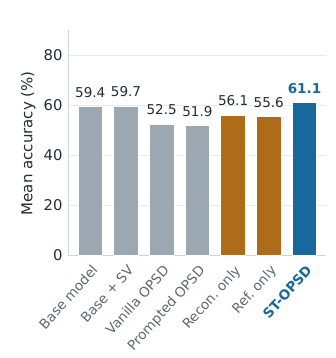}
        \caption{Method variants}
        \label{fig:ablation-variants}
    \end{subfigure}\hfill
    \begin{subfigure}[t]{0.32\textwidth}
        \centering
        \includegraphics[width=\linewidth]
            {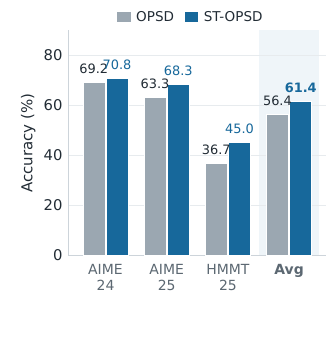}
        \caption{Omni-Math training}
        \label{fig:ablation-omnimath}
    \end{subfigure}\hfill
    \begin{subfigure}[t]{0.32\textwidth}
        \centering
        \includegraphics[width=\linewidth]
            {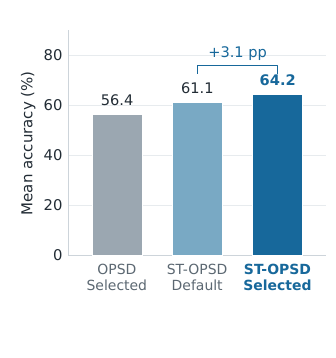}
        \caption{Combining with data selection}
        \label{fig:ablation-subset}
    \end{subfigure}

    \caption{
        Ablation studies with Qwen3-4B.
        (a) Method variants under the default training setting.
        Base + SV applies inference-time steering,
        Prompted OPSD uses teacher instructions, and
        Recon.\ only and Ref.\ only apply each steering
        direction separately.
        (b) Vanilla OPSD and ST-OPSD trained on Omni-Math.
        (c) Combining ST-OPSD with the selected subset improves
        upon the previous best ST-OPSD result, obtained with
        the default training data.
        Selected denotes examples where the teacher contributes
        new reasoning ideas.
        Panels (a) and (c) report mean accuracy across
        AIME24, AIME25, and HMMT25.
        All scores are percentages.
    }
    \label{fig:ablation}
    \vspace{-2mm}
\end{figure*}

\paragraph{Effects of steering variants.}
ST-OPSD addresses reference citation and reconsideration suppression through two teacher-side steering directions. We examine the benefit of addressing both behaviors and compare our design with direct inference-time steering and teacher prompting. In detail, we evaluate the base model, the base model with inference-time steering, Vanilla OPSD, OPSD with teacher instructions, ST-OPSD with only the reconsideration direction, and the full method. The teacher's instructions encourage reconsideration and understanding of the reference while discouraging direct reference citation, and prompt OPSD to test whether instructions can provide similar benefits by asking to encourage reconsideration and understanding of the reference directly. Finally, inference-time steering tests whether the gains can be obtained by applying steering directly to the base model during generation. \Cref{fig:ablation} reports the results under the default training setting. We have the following observations:

\noindent \textbullet \quad Reconsideration-only or reference-only steering improves upon Vanilla OPSD but remains below the base model and full ST-OPSD. Adding another direction further improves performance and brings ST-OPSD above the base model. Restoring reconsideration, therefore, partially recovers the performance lost during distillation, while correcting the reference citation provides an additional benefit. These results support addressing both teacher behaviors together.

\noindent \textbullet \quad Prompted OPSD performs worse than Vanilla OPSD. Explicitly requesting the desired behaviors, therefore, does not yield better distillation. This comparison supports the use of hidden-state interventions to obtain more effective teacher supervision.

\noindent \textbullet \quad Direct inference-time steering leaves the base model's mean accuracy nearly unchanged. ST-OPSD achieves a larger improvement, which is retained in a student who uses no steering at inference. 

\paragraph{Effects of training data.}
We further examine whether ST-OPSD retains its advantage when the training data change. We compare Vanilla OPSD and ST-OPSD when training on Omni-Math and on a selected training subset in which the teacher contributes new reasoning ideas. The latter setting tests whether steering remains beneficial when training examples are selected for this property. \cref{fig:ablation}(b) and (c) report the results, and we have the following observations.

\noindent \textbullet \quad ST-OPSD retains its advantage over Vanilla OPSD when training on Omni-Math, with improvements across all three evaluation benchmarks. This suggests that the benefit of correcting teacher supervision extends beyond our default training setting and dataset. 
    
\noindent \textbullet \quad On the selected subset, ST-OPSD still outperforms Vanilla OPSD across all three benchmarks. Vanilla OPSD remains below the base model in mean accuracy, despite using examples selected for the teacher's new reasoning ideas. This suggests that the teacher's contribution of new reasoning alone does not guarantee improved student performance. 

\noindent \textbullet \quad On the selected subset, ST-OPSD achieves the highest mean accuracy among all evaluated Qwen3-4B settings, reaching 64.2\%. It outperforms ST-OPSD trained on either the default data or Omni-Math. This suggests that selecting examples with new teacher reasoning and correcting the teacher's supervision can complement each other to further improve distillation.

\subsection{Efficiency of ST-OPSD (RQ3)}

\label{sec:efficiency}

ST-OPSD introduces hidden-state interventions during
the training stage. To evaluate the additional cost provided by ST-OPSD, we evaluate the computational time of ST-OPSD by comparing its generation and training time and token throughput with Vanilla OPSD and Purified OPSD. In detail, we compare the computation speed and cost for them using OLMo-7B and the OpenThought dataset over the 200 training steps. All runs are compared with one Nvidia H100, and \cref{tab:efficiency} presents the comparison. Overall, we have the following observation:

\begin{table}[t]
    \centering
    \small
    \setlength{\tabcolsep}{4pt}
    \begin{tabular}{lrrrr}
        \toprule
        & \multicolumn{2}{c}{Time (s/step)} 
        & \multicolumn{2}{c}{Throughput (token/s)}  \\
        \cmidrule(lr){2-3}
        \cmidrule(lr){4-5}
        Method & Gen. & Train. & Gen. & Train. \\
        \midrule
        Vanilla OPSD & 181.2 & 71.0  & 850 & 2,176 \\
        ST-OPSD      & 214.2 & 86.7  & 845 & 2,092 \\
        Purified OPSD          & 222.0 & 117.0 & 844 & 1,602 \\
        \bottomrule
    \end{tabular}
    \caption{
        Efficiency comparison on OLMo.
        Gen.\ and Train.\ denote generation and training,
        respectively.
        Lower time and higher throughput indicate
        better efficiency. The results show that ST-OPSD roughly maintains the speed.
    }
    \label{tab:efficiency}
    \vspace{-2mm}
\end{table}

\noindent \textbullet \quad ST-OPSD achieves nearly identical generation throughput to Purified OPSD and approximately 31\% higher training throughput on the same machine. These results demonstrate that ST-OPSD offers a more computationally efficient alternative to Purified OPSD since Purified OPSD requires two more forward pass during training.

\noindent \textbullet \quad Although we provide the same output budget, with teacher steering, the student in ST-OPSD tends to generate longer reasoning trajectories, while Vanilla OPSD tends to produce shorter responses during training. The resulting increase in tokens per step largely explains ST-OPSD's longer generation and training times.

\noindent \textbullet \quad Generating the trajectories for steering-vector construction takes approximately 15 minutes. Since the resulting vectors are reused throughout training, this generation cost is amortized across all training steps and adds only a small overhead relative to the distillation run.

%% file: 5conclusion.tex
\section{Conclusion}
We introduce ST-OPSD, a teacher-steering framework that mitigates reference hallucinations and reconsideration suppression in on-policy self-distillation. ST-OPSD uses two trajectory-derived steering directions to adjust the teacher's distillation targets, while the student learns through parameter updates without direct steering. Experiments on AIME and HMMT demonstrate improved accuracy over vanilla OPSD and other baselines, along with reconsideration and reference-hallucination rates restored to base-model levels. These findings suggest that shaping teacher supervision can improve distillation while preserving behaviors that support independent reasoning.

%% file: 6appendix.tex
\clearpage
\appendix

\section{Hesitation and Reference Citation Markers}

Figure~\ref{fig:hesitation-markers} lists the hesitation marker inventory, grouped into verification, backtracking, and hedging families.

\begin{figure}[htbp]
\centering
\begin{wordcard}{Hesitation Markers}
\textbf{Verification}\par
let me check / verify / double-check\par
\medskip
\textbf{Backtracking}\par
wait / actually / hold on / no / let me reconsider\par
\medskip
\textbf{Hedging}\par
maybe / perhaps / hmm
\end{wordcard}
\caption{Hesitation markers grouped into the three reconsideration families used in ST-OPSD.}
\label{fig:hesitation-markers}
\end{figure}

Figure~\ref{fig:reference-citation-markers} lists the reference-citation marker inventory. In the Answer Statements group, \textit{is/was/says} denotes three alternative forms.

\begin{figure}[htbp]
\centering
\begin{wordcard}{Reference Citation Markers}
\textbf{Example}\par
example solution, example response, example answer, sample solution, sample response\par
\medskip
\textbf{Solution}\par
original solution, provided solution, given solution, reference solution, official solution, the solution provided, the solution given\par
\medskip
\textbf{User}\par
the user said, the user says, the user provided, the user gave, the user mentioned, the user wrote, the user stated\par
\medskip
\textbf{Answer Statements}\par
example response is/was/says, example answer is/was/says, given response is/was/says, given answer is/was/says\par
\medskip
\textbf{Example Phrases}\par
According to the example, in the example
\end{wordcard}
\caption{Inventory of reference-citation markers grouped by expression type.}
\label{fig:reference-citation-markers}
\end{figure}

\section{Prompts}
\label{app:prompts}

We provide the student and teacher prompt templates used in our experiments, including the modified teacher prompt used in the Prompted OPSD ablation. In these templates, \texttt{\{question\}} denotes the problem, and \texttt{\{reference solution\}} denotes the reference solution available to the teacher.

\paragraph{Student prompt.}
The student receives only the question, without access to the reference solution. We use this prompt for both on-policy generation during training and evaluation.

\begin{promptbox}{Student prompt}
{question}
\end{promptbox}

\paragraph{Standard teacher prompt.}
The teacher receives the question together with a reference solution presented as an example response. The prompt asks the teacher to produce its own response, including its reasoning. Vanilla OPSD and ST-OPSD use the same teacher prompt. ST-OPSD additionally applies steering to the teacher's hidden states when scoring student-generated trajectories.

\begin{promptbox}{Standard teacher prompt}
{question}
This is an example for a response to the question:
{reference solution}
Now answer with a response of your own, including the thinking process.
\end{promptbox}

\paragraph{Teacher prompt for Prompted OPSD.}
To examine whether prompt instructions can encourage the desired teacher behaviors, we append two requirements to the standard teacher prompt. The first encourages the teacher to check its reasoning and reconsider possible mistakes. The second asks it to reconsider without consulting the reference and to avoid mentioning the reference in its response. This template is used for the Prompted OPSD variant in our ablation studies.

\begin{promptbox}{Teacher prompt for Prompted OPSD}
{question}
This is an example for a response to the question:
{reference solution}
Now answer with a response of your own, including the thinking process.

While reasoning, carefully double-check every step. If anything might be wrong, stop, say so, and verify or recompute it before moving on. When you pause to check or reconsider, do not look back at the example response or compare against its answer; verify by recomputing it yourself. Do not mention the example in your response.
\end{promptbox}

%% file: ref.bib
@article{yang2025qwen3,
  title={Qwen3 technical report},
  author={Yang, An and Li, Anfeng and Yang, Baosong and Zhang, Beichen and Hui, Binyuan and Zheng, Bo and Yu, Bowen and Gao, Chang and Huang, Chengen and Lv, Chenxu and others},
  journal={arXiv preprint arXiv:2505.09388},
  year={2025}
}

@article{turner2023activation,
  title={Steering language models with activation engineering},
  author={Turner, Alexander Matt and Thiergart, Lisa and Leech, Gavin and Udell, David and Vazquez, Juan J and Mini, Ulisse and MacDiarmid, Monte},
  journal={arXiv preprint arXiv:2308.10248},
  year={2023}
}

@article{li2023inference,
  title={Inference-time intervention: Eliciting truthful answers from a language model},
  author={Li, Kenneth and Patel, Oam and Vi{\'e}gas, Fernanda and Pfister, Hanspeter and Wattenberg, Martin},
  journal={Advances in Neural Information Processing Systems},
  volume={36},
  pages={41451--41530},
  year={2023}
}

@inproceedings{panickssery2024caa,
    title = "Steering Llama 2 via Contrastive Activation Addition",
    author = "Rimsky, Nina  and
      Gabrieli, Nick  and
      Schulz, Julian  and
      Tong, Meg  and
      Hubinger, Evan  and
      Turner, Alexander",
    editor = "Ku, Lun-Wei  and
      Martins, Andre  and
      Srikumar, Vivek",
    booktitle = "Proceedings of the 62nd Annual Meeting of the Association for Computational Linguistics (Volume 1: Long Papers)",
    month = aug,
    year = "2024",
    address = "Bangkok, Thailand",
    publisher = "Association for Computational Linguistics",
    url = "https://aclanthology.org/2024.acl-long.828/",
    doi = "10.18653/v1/2024.acl-long.828",
    pages = "15504--15522"
}

@article{zou2023representation,
  title={Representation engineering: A top-down approach to ai transparency},
  author={Zou, Andy and Phan, Long and Chen, Sarah and Campbell, James and Guo, Phillip and Ren, Richard and Pan, Alexander and Yin, Xuwang and Mazeika, Mantas and Dombrowski, Ann-Kathrin and others},
  journal={arXiv preprint arXiv:2310.01405},
  year={2023}
}

@inproceedings{todd2024fv,
  title={Function vectors in large language models},
  author={Todd, Eric and Li, Millicent and Sen Sharma, Arnab and Mueller, Aaron and Wallace, Byron and Bau, David},
  booktitle={International conference on learning representations},
  volume={2024},
  pages={17282--17333},
  year={2024}
}

@inproceedings{
ilharco2023editing,
title={Editing models with task arithmetic},
author={Gabriel Ilharco and Marco Tulio Ribeiro and Mitchell Wortsman and Ludwig Schmidt and Hannaneh Hajishirzi and Ali Farhadi},
booktitle={The Eleventh International Conference on Learning Representations },
year={2023},
url={https://openreview.net/forum?id=6t0Kwf8-jrj}
}

@article{imbrasaite2025reliability,
  title={Understanding (un) reliability of steering vectors in language models},
  author={Braun, Joschka and Eickhoff, Carsten and Krueger, David and Bahrainian, Seyed Ali and Krasheninnikov, Dmitrii},
  journal={arXiv preprint arXiv:2505.22637},
  year={2025}
}

@article{zhao2025saersv,
  title={Enhancing LLM Steering through Sparse Autoencoder-Based Vector Refinement},
  author={Wang, Anyi and Wu, Xuansheng and Shu, Dong and Ma, Yunpu and Liu, Ninghao},
  journal={arXiv preprint arXiv:2509.23799},
  year={2025}
}

@article{valois2025conceptors,
  title={Steering large language models using conceptors: Improving addition-based activation engineering},
  author={Postmus, Joris and Abreu, Steven},
  journal={arXiv preprint arXiv:2410.16314},
  year={2024}
}

@inproceedings{tang2025ttsv,
  title={Model Whisper: Steering Vectors Unlock Large Language Models’ Potential in Test-Time},
  author={Kang, Xinyue and Shi, Diwei and Chen, Li},
  booktitle={Proceedings of the AAAI Conference on Artificial Intelligence},
  volume={40},
  pages={31392--31400},
  year={2026}
}

@article{li2025textual,
  title={Textual steering vectors can improve visual understanding in multimodal large language models},
  author={Gan, Woody Haosheng and Fu, Deqing and Asilis, Julian and Liu, Ollie and Yogatama, Dani and Sharan, Vatsal and Jia, Robin and Neiswanger, Willie},
  journal={arXiv preprint arXiv:2505.14071},
  year={2025}
}

@article{vyas2025thinking,
  title={Understanding reasoning in thinking language models via steering vectors},
  author={Venhoff, Constantin and Arcuschin, Iv{\'a}n and Torr, Philip and Conmy, Arthur and Nanda, Neel},
  journal={arXiv preprint arXiv:2506.18167},
  year={2025}
}

@article{subramani2025feedback,
  title={Activation Steering with a Feedback Controller},
  author={Nguyen, Dung V and Vu, Hieu M and Pham, Nhi Y and Zhang, Lei and Nguyen, Tan M},
  journal={arXiv preprint arXiv:2510.04309},
  year={2025}
}

@inproceedings{
hu2022lora,
title={Lo{RA}: Low-Rank Adaptation of Large Language Models},
author={Edward J Hu and yelong shen and Phillip Wallis and Zeyuan Allen-Zhu and Yuanzhi Li and Shean Wang and Lu Wang and Weizhu Chen},
booktitle={International Conference on Learning Representations},
year={2022},
url={https://openreview.net/forum?id=nZeVKeeFYf9}
}

@misc{maa_aime,
  title        = {American Invitational Mathematics Examination (AIME)},
  author       = {{Mathematical Association of America}},
  howpublished = {\url{https://maa.org/maa-invitational-competitions/}},
  note         = {Accessed: 2026-05-03}
}

@inproceedings{agarwal2024gkd,
  title={On-policy distillation of language models: Learning from self-generated mistakes},
  author={Agarwal, Rishabh and Vieillard, Nino and Zhou, Yongchao and Stanczyk, Piotr and Ramos Garea, Sabela and Geist, Matthieu and Bachem, Olivier},
  booktitle={International Conference on Learning Representations},
  volume={2024},
  pages={21246--21263},
  year={2024}
}

@inproceedings{gu2024minillm,
  title={Minillm: Knowledge distillation of large language models},
  author={Gu, Yuxian and Dong, Li and Wei, Furu and Huang, Minlie},
  booktitle={International Conference on Learning Representations},
  volume={2024},
  pages={32694--32717},
  year={2024}
}

@inproceedings{zhao2026opsd,
  title={Self-distilled reasoner: On-policy self-distillation for large language models},
  author={Zhao, Siyan and Xie, Zhihui and Liu, Mengchen and Huang, Jing and Pang, Guan and Chen, Feiyu and Grover, Aditya},
  journal={arXiv preprint arXiv:2601.18734},
  year={2026}
}

@inproceedings{shenfeld2026sdft,
title={Self-Distillation Enables Continual Learning},
author={Idan Shenfeld and Mehul Damani and Jonas H{\"u}botter and Pulkit Agrawal},
booktitle={Forty-third International Conference on Machine Learning},
year={2026},
url={https://openreview.net/forum?id=qA6FgH0nnZ}
}

@inproceedings{hubotter2026sdpo,
 title={Reinforcement Learning via Self-Distillation},
author={Jonas H{\"u}botter and Frederike L{\"u}beck and Lejs Deen Behric and Anton Baumann and Marco Bagatella and Daniel Marta and Ido Hakimi and Idan Shenfeld and Thomas Kleine Buening and Carlos Guestrin and Andreas Krause},
booktitle={Forty-third International Conference on Machine Learning},
year={2026},
url={https://openreview.net/forum?id=QkfkxyRizZ}
}

@inproceedings{penaloza2026privileged,
title={Privileged Information Distillation for Language Models},
author={Emiliano Penaloza and Dheeraj Vattikonda and Nicolas Gontier and Alexandre Lacoste and Laurent Charlin and Massimo Caccia},
booktitle={Forty-third International Conference on Machine Learning},
year={2026},
url={https://openreview.net/forum?id=ebZcMQImhG}
}

@misc{chen2026dualopsd,
  title={Dualopsd: Adaptive privileged teachers for on-policy self-distillation},
  author={Chen, Yutong and Guo, Guangfu and Xu, Zhichao and Liu, Kunpeng},
  journal={arXiv preprint arXiv:2608.26019},
  year={2026}
}

@misc{zhang2026lopd,
  title={Latent on-policy self-distillation},
  author={Zhang, Guibin and Lyu, Jiayang and Sun, Ran and Yu, Xinlei and Zhao, Haoyu and Ren, Qibing and Yan, Shuicheng},
  journal={arXiv preprint arXiv:2608.13040},
  year={2026}
}

@inproceedings{kim2026degrade,
title={Why Does Self-Distillation (Sometimes) Degrade the Reasoning Capability of {LLM}s?},
author={Jeonghye Kim and Xufang Luo and Minbeom Kim and Sangmook Lee and Dohyung Kim and Jiwon Jeon and Dongsheng Li and Yuqing Yang},
booktitle={Third Conference on Language Modeling},
year={2026},
url={https://openreview.net/forum?id=Az7guis46K}
}

@misc{kaur2026rethinking,
  title={Rethinking On-Policy Self-Distillation for Thinking Models},
  author={Kaur, Simran and Ri, Narutatsu and He, Yinghui and Fowl, Liam and Arora, Sanjeev},
  journal={arXiv preprint arXiv:2607.05184},
  year={2026}
}

@misc{harne2026privilegedbiased,
  title={Privileged, but biased: How pi-conditioned teachers break self-distillation},
  author={Harne, Sarthak and Karkar, Chinmay and Pandya, Yash and Awadallah, Ahmed and Nambi, Akshay},
  journal={arXiv preprint arXiv:2608.04794},
  year={2026}
}

@misc{shen2026purified,
  title={Purified OPSD: On-Policy Self-Distillation Without Losing How to Think},
  author={Shen, Zhanming and Tong, Jintao and Yan, Shaotian and Shen, Chen and Chen, Hao and Ye, Wentao and Hu, Xiaomeng and Miao, Rui and Wang, Haobo and Zhao, Junbo and others},
  journal={arXiv preprint arXiv:2607.02234},
  year={2026}
}

@misc{yang2026oglssd,
  title={OGLS-SD: On-Policy Self-Distillation with Outcome-Guided Logit Steering for LLM Reasoning},
  author={Yang, Yuxiao and Wang, Xiaoyun and Zhang, Weitong},
  journal={arXiv preprint arXiv:2605.12400},
  year={2026}
}

@misc{yang2026rlsd,
  title={Self-distilled rlvr},
  author={Yang, Chenxu and Qin, Chuanyu and Si, Qingyi and Chen, Minghui and Gu, Naibin and Yao, Dingyu and Lin, Zheng and Wang, Weiping and Wang, Jiaqi and Duan, Nan},
  journal={arXiv preprint arXiv:2604.03128},
  year={2026}
}

@inproceedings{groeneveld2024olmo,
  title={OLMo: Accelerating the science of language models},
  author={Groeneveld, Dirk and Beltagy, Iz and Walsh, Evan and Bhagia, Akshita and Kinney, Rodney and Tafjord, Oyvind and Jha, Ananya and Ivison, Hamish and Magnusson, Ian and Wang, Yizhong and others},
  booktitle={Proceedings of the 62nd annual meeting of the association for computational linguistics (volume 1: long papers)},
  pages={15789--15809},
  year={2024}
}

@inproceedings{guha2026openthoughts,
  title={Openthoughts: Data recipes for reasoning models},
  author={Guha, Etash and Marten, Ryan and Keh, Sedrick and Raoof, Negin and Smyrnis, Georgios and Bansal, Hritik and Nezhurina, Marianna and Mercat, Jean and Vu, Trung and Sprague, Zayne and others},
  booktitle={International Conference on Learning Representations},
  volume={2026},
  pages={108059--108130},
  year={2026}
}

@article{balunovic2026matharena,
  title={Matharena: Evaluating llms on uncontaminated math competitions},
  author={Balunovic, Mislav and Dekoninck, Jasper and Petrov, Ivo and Jovanovi{\'c}, Nikola and Vechev, Martin},
  journal={Advances in Neural Information Processing Systems},
  volume={38},
  year={2026}
}

@inproceedings{gao2025omni,
  title={Omni-math: A universal olympiad level mathematic benchmark for large language models},
  author={Gao, Bofei and Song, Feifan and Yang, Zhe and Cai, Zefan and Miao, Yibo and Dong, Qingxiu and Li, Lei and Ma, Chenghao and Chen, Liang and Tang, Zhengyang and others},
  booktitle={International Conference on Learning Representations},
  volume={2025},
  pages={100540--100569},
  year={2025}
}

@inproceedings{kwon2023efficient,
  title={Efficient memory management for large language model serving with pagedattention},
  author={Kwon, Woosuk and Li, Zhuohan and Zhuang, Siyuan and Sheng, Ying and Zheng, Lianmin and Yu, Cody Hao and Gonzalez, Joseph and Zhang, Hao and Stoica, Ion},
  booktitle={Proceedings of the 29th symposium on operating systems principles},
  pages={611--626},
  year={2023}
}

@misc{kydlicek2025mathverify,
  author = {Kydl{\'i}{\v{c}}ek, Hynek},
  title  = {{Math-Verify}: Math Verification Library},
  year   = {2025},
  url    = {https://github.com/huggingface/Math-Verify}
}

@article{peng2026diagnosing,
  title={Diagnosing and Mitigating Thinking Collapse in On-Policy Self-Distillation},
  author={Peng, Keqin and Li, Chen and Ouyang, Yuanxin and Yuan, Yancheng and Ding, Liang},
  journal={arXiv preprint arXiv:2607.10805},
  year={2026}
}
